\documentclass[sigconf]{acmart} %,review]{acmart}

\usepackage{amsmath,amssymb,amsfonts}
\usepackage{algorithmic}
\usepackage{graphicx}
\usepackage{textcomp}
\usepackage{xcolor}
\usepackage{amsthm}
\usepackage{caption}
\usepackage{subcaption}
\usepackage{wrapfig}
\usepackage{tikz}
\usepackage{dsfont}
\usepackage{multicol}
\usepackage{multirow}
\usepackage{enumitem}
\usepackage{bm}
\usepackage{xr}
\usepackage{booktabs,colortbl,tabularx}
\usepackage{xcolor}
\usepackage{siunitx}
\usepackage{adjustbox}
\usepackage{makecell}

\AtBeginDocument{%
  }

\copyrightyear{}
\acmYear{}
\setcopyright{cc}
\setcctype{by-nc-sa}
\acmDOI{}
\acmConference[]{}
\acmBooktitle{}
\acmISBN{}
\begin{document}

\title{Few-Shot Inference of Human Perceptions of Robot Performance in Social Navigation Scenarios}

% \author{Qiping Zhang}
% \affiliation{%
%   \institution{Yale University}
%   % \city{New Haven}
%   % \state{CT}
%   % \country{USA}
% }
% \email{qiping.zhang@yale.edu}
% % \authornote{Both authors contributed equally to this research.}
% \author{}

% \author{Anonymous Author}
% \affiliation{%
%   \institution{Anonymous Institution}
% %   \city{City}
%    \country{Country}
% }
% \email{email}

% \settopmatter{authorsperrow=4}
\author{Qiping Zhang}
\affiliation{%
  \institution{Yale University}
  \city{New Haven}
  \state{CT}
  \country{USA}
}
\email{qiping.zhang@yale.edu}

\author{Nathan Tsoi}
\affiliation{%
  \institution{Yale University}
    \city{New Haven}
  \state{CT}
  \country{USA}
}
\email{nathan.tsoi@yale.edu}

\author{Mofeed Nagib}
\affiliation{%
  \institution{Yale University}
    \city{New Haven}
  \state{CT}
  \country{USA}
}
\email{mofeed.nagib@yale.edu}

\author{Hao-Tien Lewis Chiang}
\affiliation{%
  \institution{Google DeepMind}
    \city{Mountain View}
  \state{CA}
  \country{USA}
}
\email{lewispro@google.com}

\author{Marynel Vázquez}
\affiliation{%
  \institution{Yale University}
    \city{New Haven}
  \state{CT}
  \country{USA}
}
\email{marynel.vazquez@yale.edu}

\renewcommand{\shortauthors}{Qiping Zhang, Nathan Tsoi, Mofeed Nagib, Hao-Tien Lewis Chiang, Marynel Vázquez}

\begin{abstract}

Understanding how humans evaluate robot behavior during human-robot interactions is crucial for developing socially aware robots that behave according to human expectations. 
%Traditional survey methods for evaluating robot behavior often collect human data after an interaction with a robot took place, making the data susceptible to cognitive biases and resulting in sparse measurements of human perceptions. Alternatively, if human data is collected during human-robot interactions, the data collection process can be disruptive to the interaction.  Thus, 
While the traditional approach to capturing these evaluations is to conduct a user study, recent work has proposed utilizing machine learning instead. However,  existing data-driven methods require large amounts of labeled data, which limits their use in practice. To address this gap,  we propose leveraging the few-shot learning capabilities of Large Language Models (LLMs) to improve how well a robot can predict a user's perception of its performance, and study this idea experimentally in social navigation tasks. To this end, we extend the SEAN TOGETHER dataset with additional real-world human-robot navigation episodes and participant feedback. Using this augmented dataset, we evaluate the ability of several LLMs to predict human perceptions of robot performance from a small number of in-context examples, based on observed spatio-temporal cues of the robot and surrounding human motion. Our results demonstrate that LLMs can match or exceed the performance of traditional supervised learning models while requiring an order of magnitude fewer labeled instances. We further show that prediction performance can improve with more in-context examples, confirming the scalability of our approach. Additionally, we investigate what kind of sensor-based information an LLM relies on to make these inferences by conducting an ablation study on the input features considered for performance prediction. Finally, we explore the novel application of personalized examples for in-context learning, i.e., drawn from the same user being evaluated, finding that they further enhance prediction accuracy. %With more sample-efficient and higher-accuracy models for predicting human perceptions of robot behavior, t
This work paves the path to improving robot behavior in a scalable manner through user-centered feedback.

\end{abstract}



% \ccsdesc[300]{Human-centered computing~Laboratory experiments}
% \ccsdesc[300]{Human-centered computing~User studies}
% \ccsdesc[300]{Computing methodologies~Artificial intelligence}


% \includegraphics[width=\linewidth]{images/teaser_2.pdf}
% \caption{We investigate to what extent Large Language Models (LLMs) can infer human perceptions of a mobile robot in navigation scenarios (a), such as robot competence and surprising behavior (b). We frame this problem as a supervised learning problem and evaluate In-Context Learning (ICL) for few-shot predictions (c).}
% \Description{}
% \label{fig:overview}

\maketitle

\begin{figure}[t!bp]
\includegraphics[width=\linewidth]{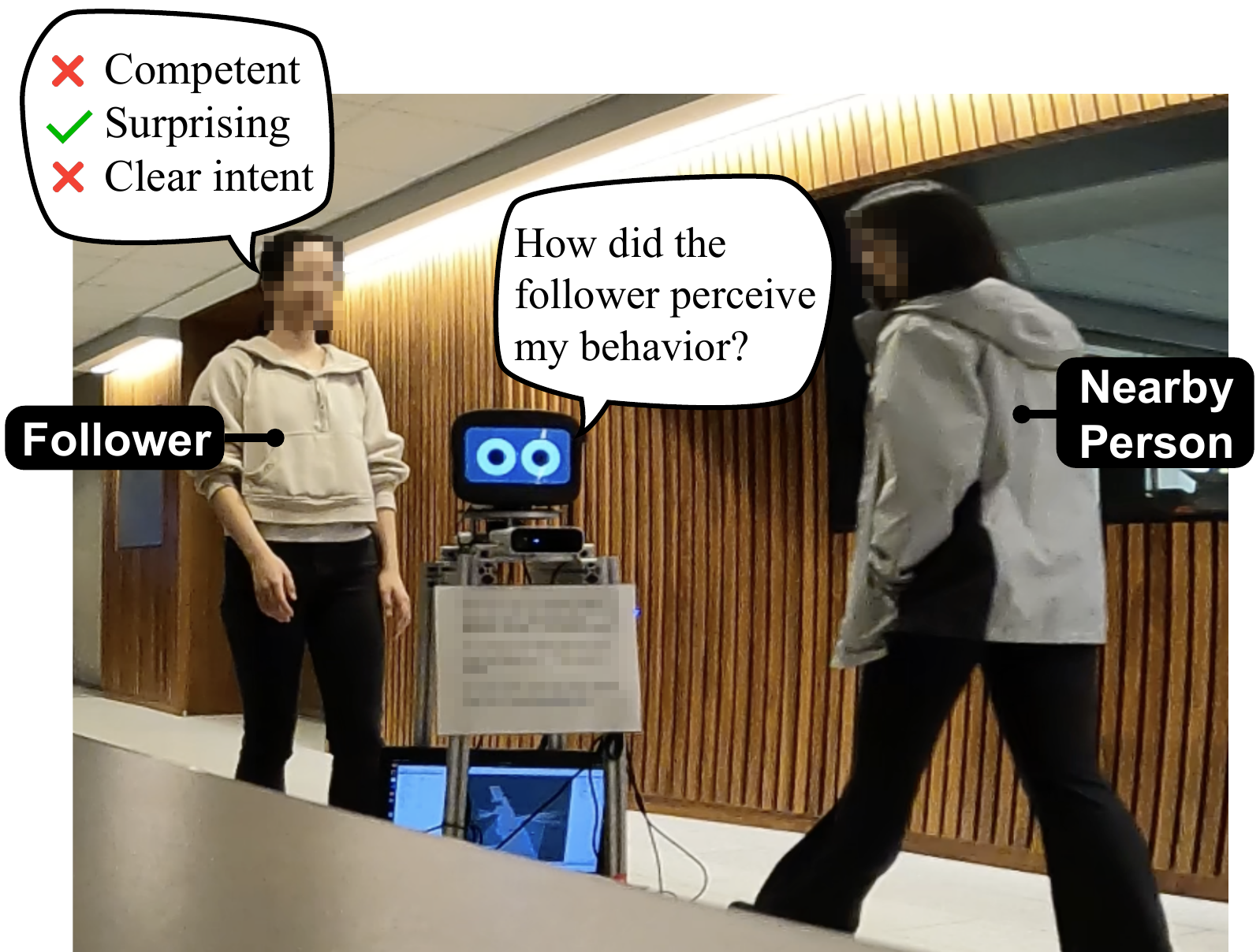}
\caption{We investigate to what extent Large Language Models (LLMs) can infer human perceptions of a mobile robot in navigation scenarios where a person -- the ``follower'' -- was guided by the robot to an indoor location. The inferences are made based on a few examples only using In-Context Learning (ICL). For each example, the input consists of sensor-based observations from the robot and the output is a binary performance level (e.g., indicating competent behavior). }
\Description{We investigate to what extent LLMs can infer human perceptions of a mobile robot using In-Context Learning (ICL)}
\label{fig:overview}
\vspace{-1em}
\end{figure}

\section{Introduction}

% \begin{figure}[t!p]
%     \centering
%     \includegraphics[width=.99\linewidth]{images/data_collection.pdf}
%     \caption{Data collection setup. Participants followed a Pioneer 3-DX mobile robot in an indoor environment with other pedestrians passing by.}
%     \label{fig:setup}
%     %\vspace{-1.5em}
% \end{figure}

% \IEEEPARstart
Inferring how humans perceive a robot’s performance is essential for designing robots that behave not only competently, but also in socially appropriate ways. These perceptions influence how people trust, collaborate with, and respond to robots in real-world settings~\cite{tan2019one,lo2019perception,pirk2022protocol,thomaz2008teachable,mitsunaga2008adapting,cui2021understanding,bera2019improving}. 
% In particular, social robot navigation tasks require robots to coordinate their movement while also aligning with human expectations about how a robot \textit{should} behave~\cite{gao2022evaluation,mavrogiannis2023core,francis2023principles}. 
% \track{This paragraph needs improvement (better flow and less repetition)... also about how robots \textit{should} behave~\cite{gao2022evaluation,mavrogiannis2023core,francis2023principles}.}
Because people’s perceptions of robot behavior are internal to the individual, they are typically measured through surveys that ask them to reflect on their experience. These evaluations tend to consider how the robot behaves in terms of different subjective factors that matter to humans, like whether the robot's actions are predictable or intentional~\cite{gao2022evaluation,mavrogiannis2023core,francis2023principles}.
However, querying people during an interaction to assess their subjective perceptions of robot behavior can be disruptive and impractical~\cite{zhang2023self,zhang2025predicting}.

Prior work proposed to use supervised learning to infer human perceptions of robots from observable interaction data~\cite{zhang2023self,zhang2025predicting}. First, survey data was collected via a user study, e.g., indicating how competent a person perceived the robot. Then, the data is used to train a learning model from scratch, such as a random forest or neural network. Once trained, the learned model can be used during interactions without having to query people via surveys again. %In that work, participants were first asked to rate robot behavior along multiple subjective dimensions, such as perceived robot competence or how surprising the behavior of the robot was to the user. Features that described the behavior of the robot and people were then used to train supervised models that predicted the users'  perceptions. 
This approach enabled more scalable robot behavior evaluation, although it required a significant data collection effort. % posing practical challenges for creating the perception models.

To make the learning approach more practical, we propose using Large Language Models (LLMs) to predict how a person perceives a robot's behavior. Because LLMs encode world knowledge and have general reasoning capabilities~\cite{brown2020language,dong2022survey}, we hypothesize that they can enable more efficient learning of human perceptions of robots than traditional supervised learning methods. Specifically, we % navigation behavior. Rather than training a new model for each target task or user, we use 
investigate using In-Context Learning (ICL) to condition a pretrained LLM on a few labeled examples and prompt it to infer a user's perception of a robot. This approach requires no retraining of the LLM, making it suitable for use in Human-Robot Interaction (HRI), where data tends to be limited and retraining of large models can easily result in overfitting. 

Our work is focused on evaluating LLMs in social robot navigation scenarios, as in Fig.~\ref{fig:overview}. %In these scenarios, interrupting a navigation task to query a user about their perceptions of the robot is heavily disruptive and, thus, learning in a sample-efficient manner is critical. 
Unfortunately, real-world social robot navigation data with human evaluations of robots is limited. Thus, we augmented an existing real-world dataset called SEAN TOGETHER~\cite{zhang2025predicting} with additional robot-guided navigation episodes.  
This augmentation expands the prior dataset from 235 interaction episodes collected from 45 participants to 404 episodes from 69 participants. For each episode, the dataset provides ground truth human perceptions of a mobile robot considering three subjective factors: whether the robot is perceived as competent, whether its behavior is surprising, and whether the robot's intentions are clear during navigation. We refer to the new dataset as the SEAN TOGETHER v2 dataset.\footnote{Link to data and code omitted for blind review.}

While prior work explored inferring human perceptions of robots with LLMs based on a high-level narrative description of an interaction~\cite{zhang2023large,claure2025-roman}, we investigate making predictions using a robot's sensor-based observations of the interaction. %The reason is that we focus our investigation on social robot navigation tasks, as in \cite{zhang2025predicting}, given  the need for user-centered evaluations of navigation behavior~\cite{francis2023principles} and the dynamic nature of these interactions. In robot navigation, human perceptions of robot behavior depend not only on the behavior of the robot of interest and the user, but also on the context of their interaction (e.g., including other nearby people). 
For example, we provide an LLM %We convert structured  data from human-robot interactions, like 
 with observed motion trajectories, each represented as a list of coordinates. %, into text for the LLM predictions. 
Using this data streamlines the application of LLMs.

Through a series of systematic experiments, our work helps us understand to what extent LLMs with ICL can infer internal human states from spatial robot data. Firstly, we analyze how LLM predictions compare to traditional supervised models in accuracy and sample efficiency. Then, we conduct an ablation study on the input features considered by an LLM, providing insights about what kind of sensor-based information it uses to infer human perceptions of a mobile robot. Finally, we investigate how tailoring demonstration examples in ICL to an individual affects the LLM's ability to infer their perceptions of robot behavior. This effort is motivated by evidence that adapting learning models in HRI to individual users can result in better prediction performance (e.g., ~\cite{rossi2017user}). To the best of our knowledge, our work is the first to explore creating personalized predictions of perceived robot performance, bringing us closer to a future where robot behavior can be evaluated at scale from a more individualized perspective than possible in prior work.

\section{Related Work}
\label{sec:related}

\noindent
\textbf{Intuitive Psychology.}
We draw inspiration from emerging studies on LLM's intuitive psychology capabilities, i.e., their ability to reason about human beliefs, goals, and social behaviors. Recent evaluations of LLMs consider varied Theory-of-Mind tasks from psychology \cite{kosinski2023theory, ullman2023large, kosinski2024evaluating, rakshit2025emotionally,  wachowiak2024large}, showing promise but also revealing brittleness under small task variations. 
%Recent work has also expanded these evaluations to multimodal settings. For instance, Schulze et al.~\cite{schulze2025visual} assess visual cognition in multimodal LLMs and frame it as a component of intuitive psychology. Their analysis builds on findings from Kosinski et al.~\cite{kosinski2024evaluating} and Ullman et al.~\cite{ullman2023large}, who suggest that powerful LLMs often struggle with theory-of-mind tasks under slight prompt perturbations. 
Evaluation suites such as CogBench \cite{coda2024cogbench} suggest that Chain-of-Thought (CoT) prompting~\cite{wei2022chain} can enhance LLM reasoning. %provide broader insights into the capabilities and limitations of LLMs in psychological inference. 
Moreover, reviews on user modeling with LLMs call for personalized interactive systems \cite{tan2023user}.

\vspace{0.5em}
\noindent
\textbf{Inferring Human Perceptions of Robots.} 
Understanding how humans perceive robot behavior is central to developing robots that are not only functional but also behave desirably. Prior work has demonstrated using subjective evaluations of robot behavior to assess robot policies~\cite{tan2019one,lo2019perception,pirk2022protocol, di2025machine, bachiller2025towards} and improve robot behavior~\cite{thomaz2008teachable,mitsunaga2008adapting,cui2021understanding,bera2019improving, song2024vlm}.

%In this work, %we study the problem of inferring human perceptions of robot performance  using data captured by a robot during a human-robot interaction. We 
We focus on predicting human perceptions that are critical in social robot navigation~\cite{gao2022evaluation,francis2023principles}: robot competence, surprisingness, and clear intent. Competence reflects the robot’s ability to perform its intended task effectively \cite{carpinella2017robotic,mavrogiannis2022social,tsoi2021approach,angelopoulos2022you}. Surprisingness captures how much a robot’s behavior deviates from user expectations~\cite{asavanant2021personal,francis2023principles,brandao2021experts}. Clear intent refers to how easily a human can infer the robot’s goal and direction of motion \cite{dragan2013legibility,dragan2015effects,sciutti2018humanizing}. These dimensions have been shown to shape people’s ability to coordinate with robots and their overall experience. Other perceptions such as discomfort~\cite{kidokoro2013will,carpinella2017robotic} and safety~\cite{akalin2022you,rubagotti2022perceived} are also relevant, but are left as future work.

Prior research showed that it is possible to use supervised learning to predict human perceptions of robots~\cite{zhang2023self, zhang2025predicting,bachiller2025towards}. In particular, we build directly on the work by Zhang et al.~\cite{zhang2025predicting}, who trained data-driven models (like a random forest model) to predict how people perceive a mobile robot during navigation. Different to prior work, though, we investigate few-shot learning, e.g., we consider learning from 4 examples versus 200+ examples as in ~\cite{zhang2025predicting}. To achieve sample-efficiency, we propose to use LLMs for the inference task. 

Other recent work in HRI explores using LLMs for zero-shot inference, e.g., to predict human trust towards a robot~\cite{zhang2023large}, identify socially-appropriate robot navigation paths~\cite{shi2025hribench}, identify robot errors~\cite{lee2025human}, predict whether robot actions are explicable or legible~\cite{verma2024theory}, and whether a robot acted fairly~\cite{claure2025-roman}. 
% Recent benchmarks in HRI \cite{shi2025hribench, lee2025human} provide valuable infrastructure for evaluating LLMs in robot-related tasks. However, these studies primarily emphasize zero-shot evaluation or vision-language settings. In contrast, this work focuses on structured textual representations of navigation behavior and explores how varying prompt composition—particularly the number and source of demonstrations—affects LLM-based perception inference.
While zero-shot prompting is practical, our results suggest that providing a few examples to LLMs can improve how well they infer human perceptions of robots.

\vspace{0.5em}
\noindent
\textbf{Few-Shot Learning with LLMs.}
In-context learning (ICL) consists of conditioning LLMs on demonstration examples at inference time, without modifying the model's parameters \cite{brown2020language,dong2022survey, mirchandani2023large, li2024llms}. This makes ICL  appealing for robotics, where adaptation to new situations and users is often required, and where full model fine-tuning — such as with LoRA \cite{hu2022lora} — can be impractical due to computational cost and latency. Thus, ICL has gained popularity for adapting robot behavior (e.g., ~\cite{di2024keypoint,wang2025inclet,yin2025context}). To our knowledge, our work is the first to use ICL to infer human perceptions of robots.

\begin{figure*}[t!p]
    \centering
    \includegraphics[width=1.0\linewidth]{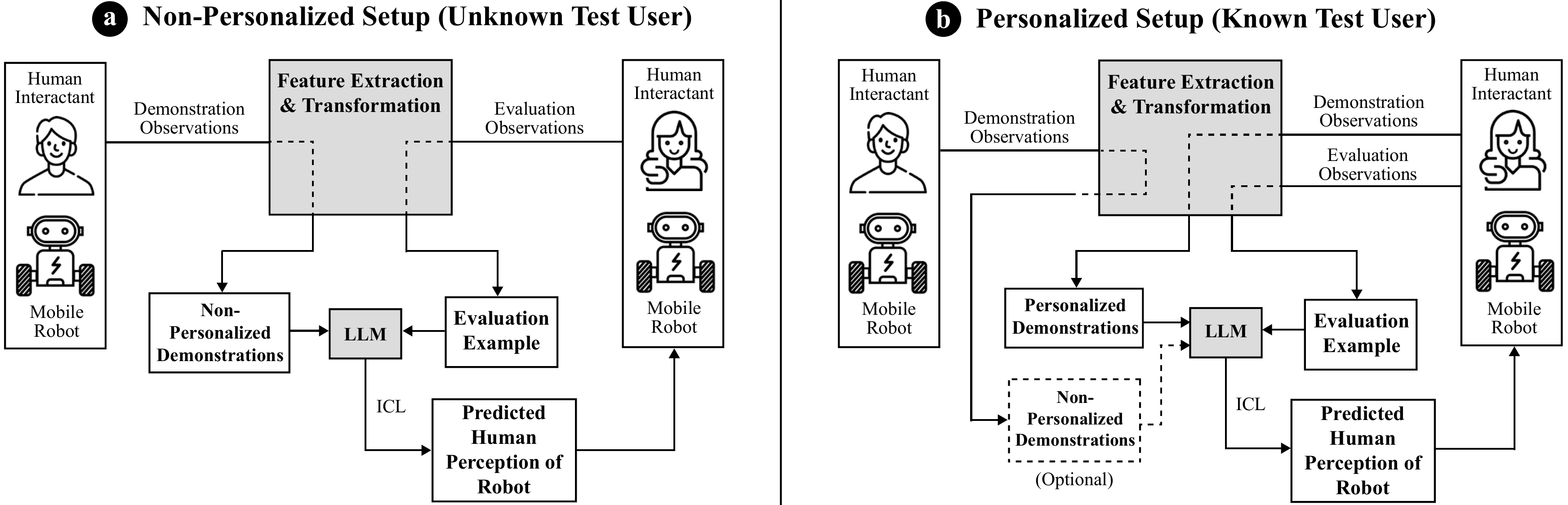}
    \caption{ICL overview: An LLM predicts a person's perception of a robot on an \emph{evaluation} example given a set of \emph{demonstrations} in the prompt. In (a),  demonstrations are gathered from interactions with users who are different from the person who generated the evaluation example. In (b), the demonstrations include examples from the same user who provided the evaluation example. }
    \label{fig:prompt_flow}
\end{figure*}

\vspace{0.5em}
\noindent
\textbf{LLM-as-a-judge.} Our work can be seen as an instance of the ``LLM as a judge'' paradigm, whereby an LLM is used to evaluate and assess the quality, relevance, or accuracy of outputs generated by other AI models. Prior work in machine learning has investigated whether LLMs with zero-shot or few-shot prompting can define rewards for Reinforcement Learning~\cite{kwon2023reward} and model user judgments about the behavior of an LLM~\cite{linunlocking,dong2024can,lau2024personalized}. % or other tasks like predicting a person's recidivism risk~\cite{halim2025let}. 
For example, Kwon et al. \cite{kwon2023reward} proposed prompting an LLM to assign reward values to an agent based on state-action trajectories from interaction scenarios. Their results demonstrate the feasibility of using LLMs to model user-aligned evaluations in games like the Ultimatum Game. %, 2-player Matrix Games, and the deal-or-no-deal negotiation task. 
Further, Dong et al.~\cite{dong2024can} proposed to personalize LLM predictions by providing a description of a persona, and Lau et al. \cite{lau2024personalized} proposed to use the ICL capabilities of transformers to dynamically adapt LLM behavior to individual preferences in simulated human populations. Inspired by this work, we  study the impact of ICL in HRI and evaluate  personalized prompting %to infer rich, perception-based judgments (e.g., surprisingness or clarity of intent) 
in physical navigation tasks.

%tasks like navigation and gridworld games. 
%Our work reinforces the central finding from this line of work:  LLMs can infer meaningful behavioral evaluations with only a few examples. However, we go further than prior work by applying this capability to richer, perception-based judgments (e.g., surprisingness or clarity of intent) in physical navigation tasks. %, rather than scalar rewards. 
 %, and investigate inference based on structured spatiotemporal features rather than high-level text summaries. 

%Our work brings together insights from intuitive psychology, few-shot learning, and human-robot interaction to explore how LLMs can function as efficient and adaptive evaluators of robot behavior from humans' perceptive.

%===============================================================================
\section{Method}
\label{sec:problem}

%Understanding how humans evaluate robot performance during navigation is central to designing robots that behave in socially appropriate ways. Prior work has demonstrated that it is possible to infer human perceptions of robot performance---such as perceived competence or surprisingness---from implicit behavioral cues during human-robot interactions~\cite{zhang2023self, zhang2025predicting}. Such approaches typically rely on supervised machine learning models that map a sequence of spatio-temporal observations from an interaction to a perception label provided by the human observer. While effective, these methods often require substantial amounts of labeled data and training effort, which limits their scalability in real-world applications.

We propose using LLMs with In-Context Learning %. ICL refers to the ability of pretrained LLMs to perform a task by conditioning on a few task demonstrations provided as part of the input prompt, without any additional parameter updates
%~\cite{dong2022survey} 
to predict human perceptions of robots. % during human-robot interactions. %Specifically, we examine whether LLMs can predict perceptions of robot performance in social robot navigation episodes, using a few demonstration examples in the LLM's prompt. 
The remaining of this section describes the ICL approach applied to a navigation scenario to facilitate the explanation; however, the same ICL formulation could be applied to other HRI interaction scenarios in the future. 

Let $\mathcal{D} = \{ (p^i,\mathbf{o}^i, y^i) \}_{i=1}^N$ be a dataset of human-robot interaction episodes, each having a finite time horizon $T$, that were collected when the robot interacted with a given person $p^i$. A sample $(p^i,\mathbf{o}^i, y^i)$ has three values: the person index $p^i \in [1,P]$, a set of robot observations $\mathbf{o}^i$, and a perception label $y^i$ provided by the person $p^i$ in relation to the robot's behavior. %The person $p^i$ interacted with the robot during the corresponding episode $i$. 
The label was collected at the end of the episode, when the person $p^i$ completed a survey to provide their momentary perceptions of the robot.  Following \cite{zhang2025predicting}, we consider these perceptions as binary labels, e.g., a $y^i$ indicates whether the robot behaved competently ($y^i = 1$) or not ($y^i = 0$) according to the person $p^i$. The observations $\mathbf{o}^i$ are gathered by the robot during the  interaction episode $i$ and, for example, include motion trajectories for the robot, the person $p^i$, and other nearby people over the time horizon $T$. Because our evaluation considers interactions where a robot guides the person $p^i$ in an indoor environment, we refer to this person as the robot's ``follower'' (Fig.~\ref{fig:overview}). %The label $y_i \in \mathcal{Y}$  corresponds to $p^i$'s perception of the robot’s performance during the interaction episode, e.g., whether the robot was perceived as competent or not. As in~\cite{zhang2025predicting}, we assume that each label $y_i$ is collected at the end of the episode and reflects the human's momentary impression of the robot's behavior. % based on their observations of the robot in the episode. %In the case of robot navigation, the observations $\mathbf{o}$ include the navigation goal, and spatial behaviors of the robot and people. 

%The goal is to approximate a function $f: \mathcal{O} \rightarrow \mathcal{Y}$ that maps the observations of the interaction gathered by the robot to a perception label. In traditional supervised learning settings, as in \cite{zhang2025predicting}, this function is learned by optimizing model parameters on a training set with \emph{demonstration} examples. Then, model performance is measured on \emph{evaluation} examples from a previously-unseen test set. In contrast, with in-context learning, we frame this task as a text-based classification problem, where information from an observation-label pair $(\mathbf{o}, y)$ and associated person index $p$ is encoded into a natural language description $s$. A prediction on a test example is made by a pretrained LLM that is conditioned on a small number of demonstration examples.

\subsection{ICL Setups}
\label{sec:ICL-setups}

Fig.~\ref{fig:prompt_flow} illustrates the ICL setups that we consider in this work, which differ in terms of how prediction performance is measured. To explain the difference, consider an LLM $\mathcal{M}$ and a previously-unseen  \emph{evaluation} example $(p^{eval}, \mathbf{o}^{eval}, y^{eval})$ from a test set generated from $\mathcal{D}$. 
%The model has to predict $y^{eval}$ given a query $q=s(\mathbf{o}^{eval})$, a string representation of the robot observations, and a collection of \emph{demonstration} examples. %and let the query $q=s(\mathbf{o}^{eval})$ be a string representation of the robot observations, which were gathered during the interaction between the robot and the person $p^{eval}$. 
The main goal of the LLM is to correctly predict the label $y^{eval}$ based on a string representation of the robot observations $\mathbf{o}^{eval}$, which we refer to as the query $q=s(\mathbf{o}^{eval})$, with $s(\cdot)$ returning the string representation. To make a prediction, the LLM is additionally provided with the following information:
\begin{equation}
    C = \{I, D_{\not=}, D_{\approx}\}
    \label{eq:llm-input}
\end{equation}
where $I$ is the  task instruction, and $D_{\not=} \in \mathcal{D}$ and $D_{\approx} \in \mathcal{D}$ are two different sets of demonstrations:
\begin{description}[style=unboxed,leftmargin=0cm,font=\normalfont\itshape,itemsep=0.5em]

\item[- Non-personalized demonstrations.] The set $D_{\not=}$  is gathered from interactions with users other than $p^{eval}$: $D_{\not=} = \big\{ s(p^i,\mathbf{o}^i, y^i) \big\}_{i = 1}^L$ where $p^i \not=p^{eval}$ and the function $s(\cdot)$ transforms the data to strings so the LLM can ingest it.

\item[- Personalized demonstrations.] The set $D_{\approx}$  is gathered from interactions with the person $p^{eval}$: $D_{\approx} = \big\{ s(p^i,\mathbf{o}^i, y^i) \big\}_{i = 1}^M$ where $p^i =p^{eval}$ but $\mathbf{o}^i \not= \mathbf{o}^{eval}$. As before, $s(\cdot)$ transforms data to strings.
\end{description}
\noindent
Then, $C$ has $K = |D_{\not=}| + |D_{\approx}|$ demonstration examples in total. Finally, the LLM  $\mathcal{M}$ makes a prediction $\hat{y}$ for the target $y^{eval}$ as:
\begin{align}
    \hat{r} &= \arg\max_{r \in \mathcal{R}} f_\mathcal{M}(r, C, q)\\
    \hat{y} &= \text{parse}(\hat{r})
    \label{eq:llm-prediction}
\end{align}
with $\hat{r} \in \mathcal{R}$ being the LLM's string response, generated in an autoregressive manner with the model's scoring function $f_\mathcal{M}$~\cite{dong2022survey}.  The prediction $\hat{y}$ is extracted from the response $\hat{r}$ via a parser. In contrast to supervised learning (as in ~\cite{zhang2025predicting}), the ICL formulation does not involve any model fine-tuning and relies solely on prompting. % the pretrained language model $\mathcal{M}$.

%We consider two setups for predicting human perceptions of robots with ICL: a \textbf{\emph{non-personalized} setup} (which induces a data split similar to the supervised learning setups from prior work~\cite{zhang2023self,zhang2025predicting}); and a \textbf{\emph{personalized} setup}, which we evaluate for the first time in the context of predicting human perceptions of robots. %To explain the difference, assume that a machine learning model is going to make a prediction on a previously-unseen \emph{evaluation} example $(p^{eval}, \mathbf{o}^{eval}, y^{eval})$ from a test set. The model has to predict $y^{eval}$ given $\mathbf{o}^{eval}$ and a collection of \emph{demonstration} examples. In the non-personalized setup, there are no demonstration examples in the training data that were collected from interacting with the person $p^{eval}$.  Thus, the non-personalized setup corresponds to making predictions for an \emph{unknown} test user, as illustrated in Figure~\ref{fig:models}(a). However, in the personalized setup, there are demonstration examples in the training data that were collected with the person $p^{eval}$ -- although an evaluation example from the test set is never included in the train data to avoid data leakage. The personalized setup thus corresponds to making predictions for a \emph{known} test user, as illustrated in Figure~\ref{fig:models}(b).

In the \textit{\textbf{non-personalized ICL setup}}, there are no demonstration examples that were collected from interacting with the person $p^{eval}$. This corresponds to making inferences when the set of personalized demonstration examples \emph{is empty}, $C = \{I, D_{\not=}, \{\}\}$. Thus, the non-personalized setup can be seen as making predictions for an \emph{unknown} test user, as in Fig.~\ref{fig:prompt_flow}(a). This setup induces a data split similar to the supervised learning setups from prior work~\cite{zhang2023self,zhang2025predicting}.

Conversely, the \textit{\textbf{personalized ICL setup}} corresponds to making predictions when the set of personalized demonstration examples \emph{is not empty}. This can be seen as making predictions for a \emph{known} test user, as illustrated in Fig.~\ref{fig:prompt_flow}(b).  We study two cases for the latter setup: in one case,  only personalized examples are provided, so $C = \{I, \{\}, D_{\approx}\}$; in another case, both non-personalized and personalized examples are provided, so $C = \{I, D_{\not=}, D_{\approx}\}$. Fig.~\ref{fig:prompt} illustrates the prompt structure for the non-personalized and personalized setups.

\subsection{Observation Space}
\label{sec:observations}

Based on findings from prior work~\cite{zhang2025predicting}, we utilize spatial behavior features for predicting perceived robot performance in social navigation scenarios. %, where a person follows the robot to a goal location in an indoor environment. 
These features can be computed by mobile robots  using off-the-shelf approaches for people tracking (e.g., with Kinect sensors~\cite{shotton2011real}) and for robot localization~\cite{grisetti2007improved}.

For a given example $(p, \mathbf{o}, y)$, the observations $\mathbf{o}$ provide a temporally grounded and robot-centric view of the navigation scene, encoding how the robot, the participant, and others in the environment move over time. % with respect to the robot’s starting configuration. 
The observations $\mathbf{o}$ span an 8-second time horizon and are represented in a coordinate frame centered on the robot’s pose at the initial timestep ($t=0$). Temporally-varying data is sampled at 1 Hz. %All data are represented in a coordinate frame centered on the robot’s pose at the initial timestep ($t=0$), providing an egocentric view of the scene and eliminating dependence on global location or orientation.
The observations are:
\begin{itemize}[wide, labelindent=0em,leftmargin=0em,itemsep=0em,topsep=0em]
    \item[--] \textit{Goal Pose:} The 2D position of the robot’s navigation goal relative to the robot at $t=0$.
    \item[--] \textit{Robot Trajectory:} The robot’s 2D position and orientation  %, represented as a \((\cos\theta, \sin\theta)\) pair, 
    at each timestep of the time horizon.
    \item[--] \textit{Follower Trajectory:} The 2D position %\((x, y)\) 
    and orientation of the person following the robot at each timestep of the time horizon.
    \item[--] \textit{Nearby Pedestrians:} The 2D positions and orientations of other pedestrians at each timestep of the time horizon. We consider only observed people within a 7.2-meter radius of the robot, which corresponds to the robot's public space per Hall's proxemic zones~\cite{hall1966hidden}.
\end{itemize}

All 2D positions are encoded as \((x, y)\) locations, and the orientations $\theta$ are encoded as  \((\cos\theta, \sin\theta)\). The cos-sin encoding is standard practice to ensure continuity for  learning algorithms~\cite{wang2017deepvo,swofford2020improving,zhang2023self}. Although it is not as critical for LLMs, it helps  supervised learning models, which we compare against in our evaluation. 

\begin{figure}[t!]
    \centering
    \includegraphics[width=.99\linewidth]{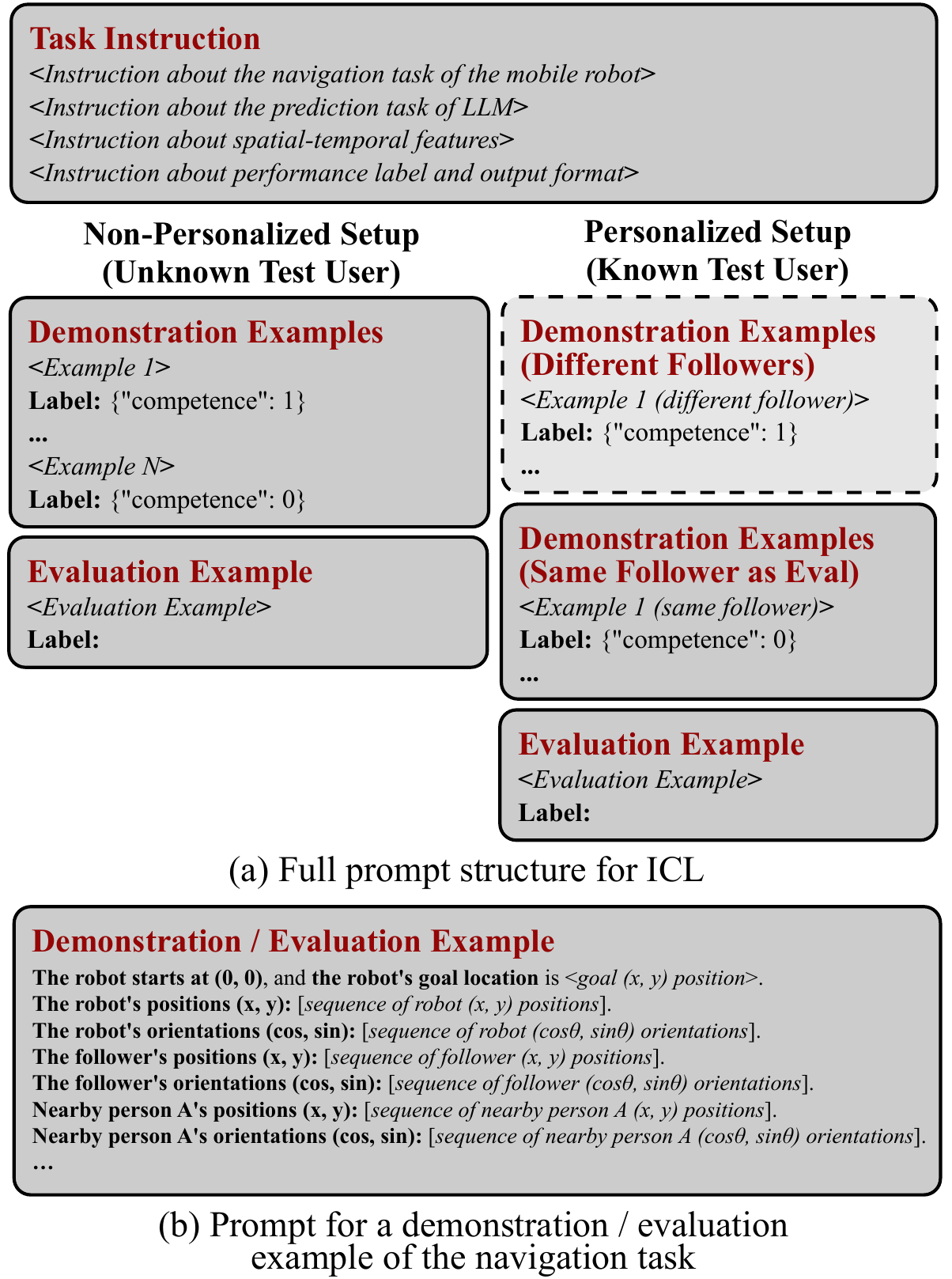}    
    \vspace{-.5em}
    \caption{Prompt structure (a), including the structure for an example (b). The LLM is asked to predict robot competence.}
    \label{fig:prompt}
    \vspace{-1em}
\end{figure}

 Fig.~\ref{fig:prompt}(b) illustrates how the observation features are included in the LLM's prompt. When a particular person is not detected in a timestep, their position and orientation are indicated as ``unknown''.

% To interface with large language models, we convert each data example into a structured natural language description. These descriptions specify the robot’s goal location, followed by sequences of 2D positions and orientations—represented as \((x, y)\) and \((\cos\theta, \sin\theta)\) pairs—for the robot, the follower, and each nearby pedestrian across the 8 timesteps. When a nearby pedestrian is not detected by the robot's onboard sensors at a given timestep, their position and orientation are marked as “unknown” in the text. This text representation structure is consistently used in our in-context learning setup, where both demonstration and query examples share the same format.

\section{Experimental Setup}
\label{sec:experiments}

We evaluate ICL for predicting human perceptions of robot performance using a real-world HRI dataset, which is an augmentation of an existing dataset. Zhang et al.~\cite{zhang2025predicting} contributed the \textsc{SEAN TOGETHER} dataset, which provides short episodes of human-robot interaction during a social navigation task in semi-public university environments. 
%using an augmented version of the real-world portion of the \textsc{SEAN TOGETHER} dataset~\cite{zhang2023self}. This dataset captures short episodes of human-robot interaction during a social navigation task in semi-public university environments. 
Each episode contained observations of the interaction captured by the robot and corresponding human perceptions. We expanded the dataset with approval of our local Review Board using the same protocol and mobile robot. This increased the number of participants from $45$ people to   $69$ people, totaling $404$ labeled interaction episodes. We named the augmented dataset SEAN TOGETHER v2. See the supplementary video for example episodes.

\vspace{0.5em}
\noindent
\textbf{Data Collection Protocol.} The  robot, which can be seen in Fig.~\ref{fig:overview}, was built on a Pioneer 3-DX base. It was equipped with two Kinect sensors (one looking forward and one backwards), and had a screen-based face. It navigated autonomously through two public indoor spaces on a university campus. One space was a pedestrian tunnel; the other was a building entrance corridor. 

The participants were not pre-recruited; rather, they engaged with the robot opportunistically. As pedestrians encountered the robot at the university, experimenters invited them to briefly follow the robot to a nearby goal marked on the ground. The robot periodically stopped during navigation and prompted the participants to evaluate its behavior using a mobile interface.  

The robot's high-level behavior was implemented as in ~\cite{zhang2025predicting}, where it either moved efficiently toward the destination (\textit{Nav-Stack} behavior), spinned in place to appear confused (\textit{Spinning} behavior), and moved away from the goal (\textit{Wrong-Way} behavior). The robot switched between the high-level behaviors to maintain a consistent rate of sub-optimal behavior. The behaviors were designed to elicit both
positive and negative views of the robot while also avoiding participant boredom or significant confusion.% which can induce various reactions of human interactants and their perceptions of robot performance.

Each behavior was executed for a fixed duration of 20-40 secs. Shortly before or after a behavior change, there was a pause in which participants rated the robot's recent performance.  They answered 3 questions on a 5-point Likert  format:  \textit{``How competent was the robot at navigating?''} (\textit{\textbf{Competence}} performance dimension); \textit{``How surprising was the robot's navigation behavior''} (\textit{\textbf{Surprise}}); and \textit{``How clear were the robot's intentions during navigation?''} (\textit{\textbf{Intention}}).
%
%An interaction episode tended to last 40 secs. 
As in \cite{zhang2025predicting}, we inferred human perceptions based on an 8-second observation window preceding the participant’s response. 

\vspace{0.5em}
\noindent
\textbf{Performance Labels.} 
Zhang et al.~\cite{zhang2025predicting} showed that predicting human perceptions of robots in a 5-point scale is very difficult even for humans. Thus, we created the performance labels for our experiments by binarizing the human ratings provided by the participants during data collection. Specifically, we excluded neutral-labeled examples, which were rare, and mapped the remaining responses %as follows: For the \textit{Competence} and \textit{Intention} dimensions, 
by converting 
ratings of 4 or 5 to positive labels, and ratings of 1 or 2 to negative labels. %For  \textit{Surprise}, ratings of 4 or 5 indicated the robot was perceived as surprising so they were positive labels. Meanwhile,  ratings of 1 or 2 indicated it was perceived as unsurprising, so they were negative labels. 
In total, we had 363  episodes for the prediction of \textit{Competence}, 351 for \textit{Surprise}, and 375 for \textit{Intention}.

\vspace{0.5em}
\noindent
\textbf{Evaluation Procedure.} Each perception dimension (\textit{Competence}, \textit{Surprise}, and \textit{Intention})  is a separate classification task. For each  dimension, we partitioned the participants into  disjoint sets: 40\% for testing, 40\% for training, and 20\% for validation. The splits were fixed across all experiments and conditions. To ensure that each participant in the test set contributed usable evaluation data, we only included participants who have at least one example with a positive label, one with a negative label, and at least four additional examples for personalization analysis. From each of the test participants, we randomly selected one positive and one negative example as the evaluation examples, resulting in a balanced evaluation set. % composed of user-specific examples, and this set 
%which was constant across experimental runs. 
Thus, we measured performance with classification \textit{accuracy}.

All experiments are repeated 25 times because we use randomized demonstration sampling, whereby the demonstration examples are randomly chosen from the training data. For each of the 25 runs, all the models utilize the same demonstration and evaluation examples, and make predictions based on the same feature representation for the observations (as in Sec.~\ref{sec:observations}) to ensure a fair evaluation.

\section{Evaluation}
\label{sec:evaluation}

% While ICL has shown promise in other applications, it is important to understand at a fundamental level whether it can serve to predict human perceptions of robots in HRI. Thus, 
We systematically investigated four research questions (RQs) %in  the navigation scenario %. Each RQ considered one of the ICL setups described previously in Sec.~\ref{sec:ICL-setups} and a 
using a limited number of demonstrations that ranged from $K=4$ to $K=64$.\footnote{Because LLMs can be biased by the distribution of target labels in the demonstration examples for ICL~\cite{wang2023large}, we always set $K$ to be a power of $2$ so that we could balance the number of positive and negative examples whenever possible.} We used linear mixed model analyses estimated with REstricted Maximum Likelihood (REML)~\cite{patterson1975maximum,stroup2012generalized} to evaluate accuracy for each performance dimension. The analyses considered Run ID as a random effect because we repeated the experiments 25 times with varying demonstrations. The independent variables varied per RQ. %, as detailed next.  %Our dependent variables were the accuracy of the models' predictions with regards to the ground-truth labels of the test examples. 

\definecolor{mplBlue}{rgb}{0.7176, 0.8431, 0.9490}
\definecolor{mplOrange}{rgb}{0.9843, 0.8902, 0.8353}
\definecolor{mplGreen}{rgb}{0.8000, 0.9250, 0.8000}

\definecolor{mplGreenLight}{rgb}{0.90, 0.97, 0.90}

% Medium green (for 2nd place) – your base pastel green
\definecolor{mplGreenMedium}{rgb}{0.75, 0.90, 0.75}

% Darker green (for 1st place) – clearly deeper but not blocking text
\definecolor{mplGreenDark}{rgb}{0.55, 0.75, 0.55}

\begin{table*}[t!p]
\centering
\begin{minipage}[t]{0.63\textwidth}
    \centering
    \fontsize{8.5pt}{10pt}
    \selectfont
    \captionof{table}{Results for RQ1. Average accuracy ($\pm \text{ std. err.}$) of LLMs with ICL and Random Forest (RF) over 25 repetitions. CoT stands for Chain-of-Thought prompting. The \colorbox{mplGreenDark}{Best},  \colorbox{mplGreenMedium}{Second}, and \colorbox{mplGreenLight}{Third} average results are highlighted.}
    \vspace{-0.1cm}
    \begin{tabular}{c|c|c|c|c|c}
      \toprule[1.5pt]
     \textbf{Model} & \textbf{CoT} & \makecell{\textbf{\#~Demo.} \\ \textbf{Examples ($K$)}} & \textbf{Competence} & \textbf{Surprise} & \textbf{Intention}\\
     \midrule
     \addlinespace[0.4em]
     Gemini 2.0 Flash & No & 4 & \cellcolor{mplGreenLight} $0.67 \pm 0.01$ & \cellcolor{mplGreenDark} $0.65 \pm 0.01$ & $0.65 \pm  0.02$\\
     Gemini 2.0 Flash & Yes & 4 & \cellcolor{mplGreenDark} $0.72 \pm 0.01$ & \cellcolor{mplGreenMedium} $0.64 \pm 0.01$ & \cellcolor{mplGreenDark} $0.69 \pm  0.01$ \\
     GPT 4.1 mini & No & 4 & $0.67 \pm 0.01$ & \cellcolor{mplGreenLight} $0.64 \pm 0.01$ & \cellcolor{mplGreenLight} $0.65 \pm 0.01$\\
     GPT 4.1 mini & Yes & 4 & \cellcolor{mplGreenMedium} $0.69 \pm 0.01$ & $0.64 \pm 0.01$ & \cellcolor{mplGreenMedium} $0.68 \pm 0.01$\\
     Llama 3.2 90B & No & 4 & $0.57 \pm 0.01$ & $0.55 \pm 0.01$ & $0.51 \pm 0.01$\\
     Llama 3.2 90B & Yes & 4 & $0.61 \pm 0.01$ & $0.52 \pm 0.01$ & $0.57 \pm 0.01$\\
     RF & / & 4 & $0.53 \pm 0.02$ & $0.57 \pm 0.01$ & $0.49 \pm 0.02$ \\
     \midrule
     Gemini 2.0 Flash & No & 64 & \cellcolor{mplGreenMedium} $0.72 \pm 0.01$ & \cellcolor{mplGreenDark} $0.70 \pm 0.01$ & \cellcolor{mplGreenDark} $0.67 \pm 0.01$\\
     Gemini 2.0 Flash & Yes & 64 & \cellcolor{mplGreenDark} $0.73 \pm 0.01$ & \cellcolor{mplGreenLight} $0.67 \pm 0.01$ & \cellcolor{mplGreenMedium} $0.67 \pm  0.01$\\
     GPT 4.1 mini & No & 64 & \cellcolor{mplGreenLight} $0.71 \pm 0.01$ & $0.67 \pm 0.01$ & \cellcolor{mplGreenLight} $0.65 \pm 0.01$\\
     GPT 4.1 mini & Yes & 64 & $0.70 \pm 0.01$ & $0.66 \pm 0.01$ & $0.64 \pm  0.01$\\
     Llama 3.2 90B & No & 64 & $0.53 \pm 0.01$ & $0.59 \pm 0.01$ & $0.45 \pm 0.01$\\
     Llama 3.2 90B & Yes & 64 & $0.51 \pm 0.01$ & $0.48 \pm 0.01$ & $0.51 \pm 0.01$\\
     RF & / & 64 & $0.66 \pm 0.01$ & \cellcolor{mplGreenMedium} $0.69 \pm 0.01$ & $0.61 \pm 0.02$ \\
     \bottomrule[1.25pt]
    \end{tabular}
    \label{tab:rq1}
\end{minipage}%
\hfill
\begin{minipage}[t]{0.35\textwidth}
    \centering
    %\vspace{0.01ex}
    \vspace{-1em}
    \includegraphics[width=1\linewidth]{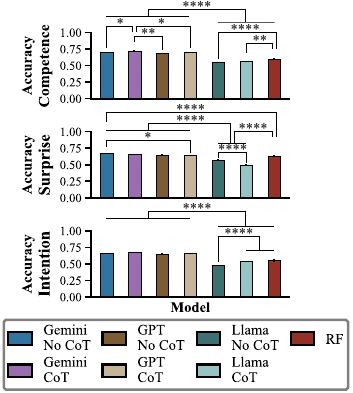}
    \vspace{-2em}
    \captionof{figure}{Model accuracy for RQ1. (****), (**), and (*) denote $p < 0.0001$, $p < 0.01$, and $p < 0.05$. Error bars are std. err. and are small.}
    \label{fig:models}
    \vspace{-0.5em}
\end{minipage}
\end{table*}

%\vspace{-1em}
\subsection{Non-Personalized ICL with Different LLMs}

Our first research question was: \\[0.3em]
\noindent
\textit{\textbf{RQ1:} Do LLMs with ICL result in more accurate, non-personalized predictions of human perceptions of robot performance in few-shot learning scenarios than more traditional supervised learning?}\\[0.3em]
In prior work~\cite{zhang2025predicting}, a Random Forest (RF) supervised learning model provided state of the art performance for predicting human perceptions of a guide robot. Thus, we compared the RF model in the the SEAN TOGETHER v2 dataset against  several LLMs: Gemini 2.0 Flash; GPT 4.1 mini; and Llama 3.2 90B. The first two LLMs are closed-source models, while the latter one is open-source. Because part of our motivation for these models is making inferences \emph{during} interactions, we limited the set of LLMs to  ``non-reasoning'' models that can produce predictions relatively quickly on the cloud, e.g.,  hundreds of tokens per second, by simply predicting one token at a time in an auto-regressive fashion. For each LLM model, we also considered two prompting strategies: one with Chain-of-Thought (CoT) reasoning \cite{wei2022chain} that asked the model to ``Do it step by step and explain your answer''; the other involved no CoT reasoning. 
% using a scoring function (as in eq.~(\ref{eq:llm-prediction})). %, rather than breaking down a problem into smaller steps and revising their thinking. 

    While ICL provides the demonstration examples to the LLMs via their prompt, the same demonstrations are used to train the RF  from scratch. We thus hypothesized that with fewer examples, the LLMs would do better than the RF. To test this idea, we compared results in two scenarios:  having few demonstrations with $K=|D_{\not=}|=4$; and having a larger number with $K=|D_{\not=}|=64$,  which approached the limit of the context window for  Llama given our prompt (Fig.~\ref{fig:prompt}). %In each case, we examine how well the learned function $f$ (or in the ICL case, $\hat{y}$ via $f_{\mathcal{M}}$) generalizes to unseen interaction episodes. This comparison is made using consistent input features and evaluation metrics, enabling us to measure the sample efficiency of LLM-based inference.

%To understand the sample efficiency of LLMs in inferring human perceptions of robot performance, we first evaluated how well LLMs could perform on this prediction problem with 4 and 64 examples used for demonstration in ICL, which are the minimum and maximum number of examples that we consider for unknown test users in our dataset, respectively. To validate the generalizability of our method, we evaluated on three different foundation models: \textit{Gemini 2.0 Flash}, \textit{GPT 4.1-mini}, and \textit{Llama 3.2 90B}. For each LLM model, we also considered two prompting strategies: one with Chain-of-Thought (CoT) reasoning \cite{wei2022chain} that asks the model to ``Do it step by step and explain your answer'', while the other involves no CoT. %We also compared with the most sample-efficient and accurate model on the SEAN TOGETHER Dataset \cite{zhang2025predicting}, which is a Random Forest (RF) classifier.

\vspace{0.5em}
\noindent
\textbf{Results.} Table~\ref{tab:rq1} shows prediction accuracy on 25 runs with $K=4$ and $K=64$ demonstration examples. %In general, more demonstrations tended to result in higher average accuracy. % although this was not always true for the LLMs.
%
%For all performance dimensions, an LLM with ICL resulted in the best average performance. %Moreover, the Gemini and GPT models yielded close accuracy, showing the generalizability of our ICL approach. 
The best result for the RF was on Surprise with $K=64$, where it provided close performance to Gemini; otherwise, the RF model underperformed Gemini and GPT.

We analyzed the accuracy results with linear mixed models, one per performance dimension. Each analysis considered Run ID as a random effect, and Number of Demonstrations ($K=4$ and $K=64$) and Model (7 levels, each row of Table~\ref{tab:rq1}) as main effects. Also, the analysis considered the interaction between the main effects. %We performed post-hoc tests when appropriate. 
 
The analysis indicated that the Number of Demonstrations ($K$)  had a significant effect on the accuracy for \textit{Competence} ($F(1, 312) = 5.94$, $p = 0.0154$) and \textit{Surprise} ($F(1, 312) = 53.38$, $p < 0.0001$). For \textit{Competence}, a post-hoc Student's t-test showed that  64 demonstrations ($M = 0.65$, $SE = 0.007$) led to significantly higher accuracy than 4 demonstrations ($M = 0.64$, $SE = 0.006$) -- although the average difference was close enough that it lacked functional meaning. A more pronounced significant difference was obtained for \textit{Surprise}, where $K=64$ led to an average accuracy of $M = 0.64$ ($SE = 0.01$), and $K=4$ led to $M = 0.60$ ($SE = 0.01$). We attribute the limited effect of the $K$ demonstrations on accuracy (considering several LLMs) to the challenge of processing long context windows~\cite{li2024long}. For example, for Gemini No CoT with $K=4$, the context window had about $3,000$ tokens, while $K=64$ led to about $36,000$ tokens.

Model had a significant effect on the prediction accuracy  %, as shown in Fig.~\ref{fig:models}. 
($p < 0.0001$ for all performance dimensions). %for Competence ($F(6, 312) = 120.46$; $p < 0.0001$), Surprise ($F(6, 312) = 91.48$; $p < 0.0001$), and Intention ($F(6, 312) = 94.25$; $p < 0.0001$). 
Fig.~\ref{fig:models} shows significant pairwise differences with Tukey HSD post-hoc tests. In general, the Gemini and GPT models resulted in significantly higher accuracy than Llama and RF. Although there were some significant differences in accuracy between the Gemini and GPT models, differences were small, showing the generalizability of our ICL approach. 

Lastly, we found that the interaction between %the Number of Demonstrations ($K$) 
$K$ and Model had a significant effect on  accuracy, with  $p < 0.0001$ for all performance dimensions. %for Competence ($F(6, 312) = 27.85$; $p < 0.0001$), Surprise ($F(6, 312) = 15.31$; $p < 0.0001$), and Intention ($F(6, 312) = 16.05$; $p < 0.0001$). 
For $K = 4$, the Gemini and GPT models resulted in significantly higher accuracy than RF and Llama in all dimensions. 
For $K=64$, the post-hoc interaction tests showed that the Llama models resulted in significantly lower accuracy, but other results varied by dimension: for \textit{Competence}, Gemini CoT and No CoT led to significantly higher accuracy than the other models, except for the GPT models; 
for \textit{Surprise}, there was no significant difference for the Gemini, GPT, and RF models; %Gemini No CoT and RF led to significantly higher accuracy than Llama, but not the GPT models nor Gemini CoT; 
and for \textit{Intention}, Gemini No CoT had significantly higher accuracy than RF. 
In addition, for \textit{Competence}, Gemini CoT with only $K=4$ led to significantly higher accuracy than RF and Llama with $K = 64$, while for \textit{Intention}, Gemini CoT and GPT CoT with only $K = 4$ led to significantly higher accuracy than RF and Llama with $K = 64$.

\subsection{The Value of Spatial Observations for ICL}

Our second research question was:\\[0.3em]
\noindent
\textit{\textbf{RQ2:} Which spatial observations drove ICL performance with limited demonstrations ($K=4$)?}\\[0.3em]
\noindent
We compared making non-personalized ICL predictions in a few-shot ICL scenario utilizing different types of observations: 1) the goal and robot trajectory only; 2) the goal, robot trajectory, and follower trajectory; and 3) the goal, robot trajectory,  follower trajectory, and other pedestrian trajectories, i.e., all observations in Sec.~\ref{sec:observations}. Thus, this RQ  served as a feature ablation for ICL. 

We limited our evaluation (and the following RQs) to Gemini 2.0 Flash with no CoT given our prior results, which showed strong performance for this model. Focusing on one model also helped reduce the cost of experiments and their carbon footprint~\cite{faizllmcarbon}. 

\begin{figure}[tb!p]
    \centering
    \includegraphics[width=.95\linewidth]{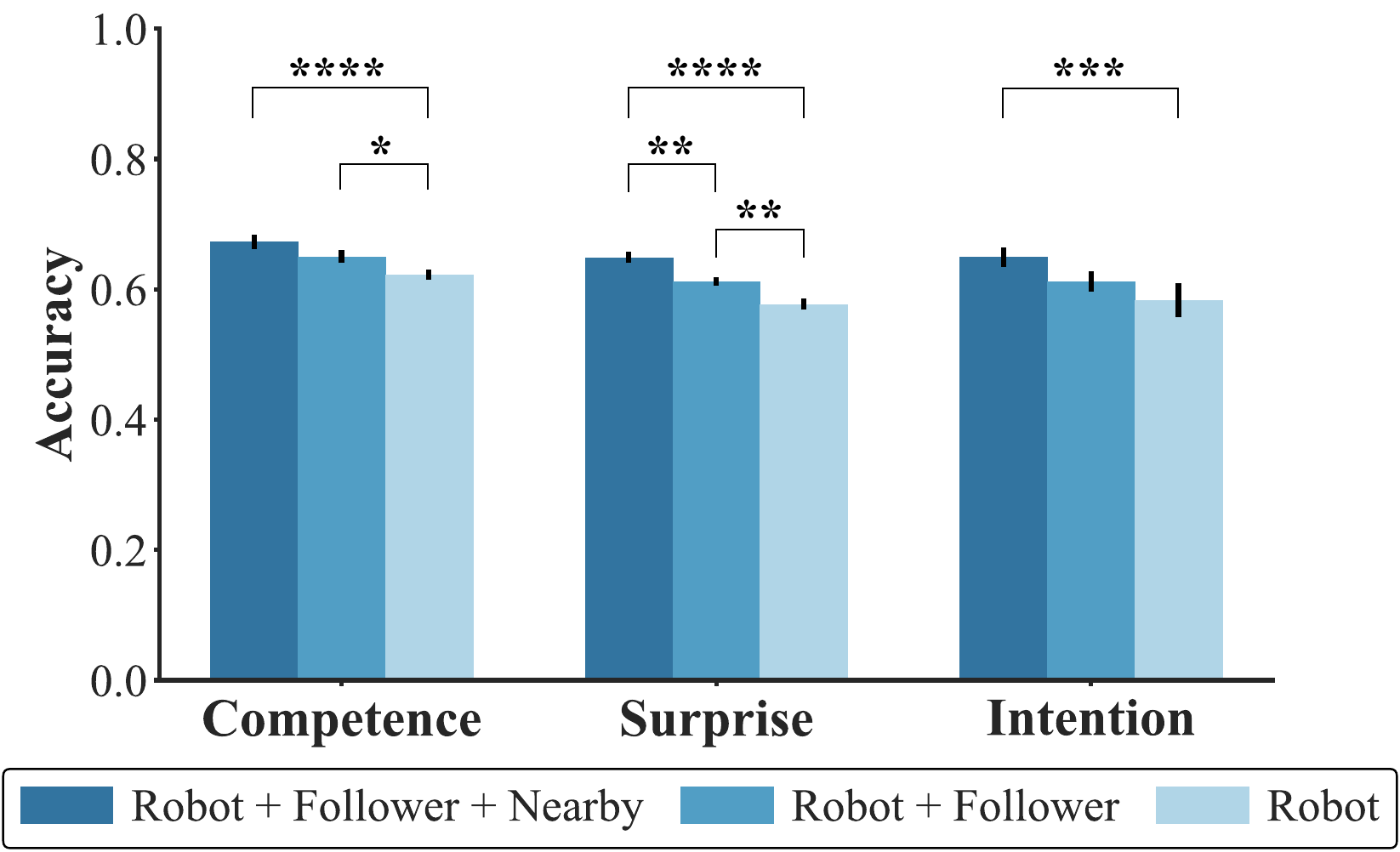}
    \caption{Results for RQ2. Average accuracy for Gemini 2.0 Flash No CoT with $K=4$. The model always takes as input the goal location, but the other spatial observations are ablated. Error bars are std. err. The symbols (****), (***), (**), and (*) denote $p < 0.0001$, $p < 0.001$, $p < 0.01$, and $p < 0.05$.}
    \label{fig:RQ2}
    \vspace{-1em}
\end{figure}

\vspace{0.5em}
\noindent
\textbf{Results.} Fig.~\ref{fig:RQ2} shows prediction accuracy. For each performance dimension, we fit a linear mixed model on accuracy considering the Set of Observations provided to Gemini as main effect, and Run ID as random effect. The Set of Observations had a significant effect on \textit{Competence} ($p < 0.0001$), \textit{Surprise} ($p < 0.0001$), and \textit{Intention} ($p = 0.0007$). Due to limited space, we summarize significant pairwise differences from Tukey HSD post-hoc tests in Fig.~\ref{fig:RQ2}. The results show that the LLM's performance was not only due to using information about the robot trajectory and goal, but also to using pedestrian observations (including observations of the follower).

\subsection{Increasing ICL Demonstrations}

Our third research question was:\\[0.3em]
\noindent
\textit{\textbf{RQ3:} How does ICL performance vary with an increasing number of demonstration examples ($K$) in the non-personalized  setup?}\\[0.3em]
\noindent
We analyzed in more detail the impact of  $K$ on the ICL predictions for Gemini 2.0 Flash with no CoT, which provided good performance in RQ1. 
We considered $K \in \{4, 8, 16, 32, 64\}$, and two supervised learning baselines trained from scratch: RF (as in RQ1), and a neural network  with a GRU architecture~\cite{chung2014empirical} (as in~\cite{zhang2023self}). Also, we considered two other baseline models that required no training. First, we compared results with a weighted random sampling model (WR) that predicted a label by sampling from the distribution of targets in the  demonstrations. This helped understand the complexity of the prediction problem. Further, we evaluated 
Gemini 2.0 Flash in a zero-shot prediction scenario, where the task instruction was the same as for ICL but no demonstrations were provided to the LLM (thus $K=0$). %, i.e., the sets $D_{\not=} = \{\}$ and $D_{\approx} = \{\}$ were empty in eq.~(\ref{eq:llm-input}). 
This helped gauge how much the demonstrations contributed to the LLM's performance given its world knowledge. % in comparison to the world knowledge encoded in the LLM.  %Using the prompt structure shown in Fig.~\ref{fig:prompt_flow}, we incrementally increase $k$ while keeping other components of the prompt fixed. This evaluation tests the scaling behavior of in-context learning and reveals whether LLMs benefit from additional in-context examples when predicting human perceptions of robot performance.

\begin{figure}[t!]
    \centering
    \includegraphics[width=.95\linewidth]{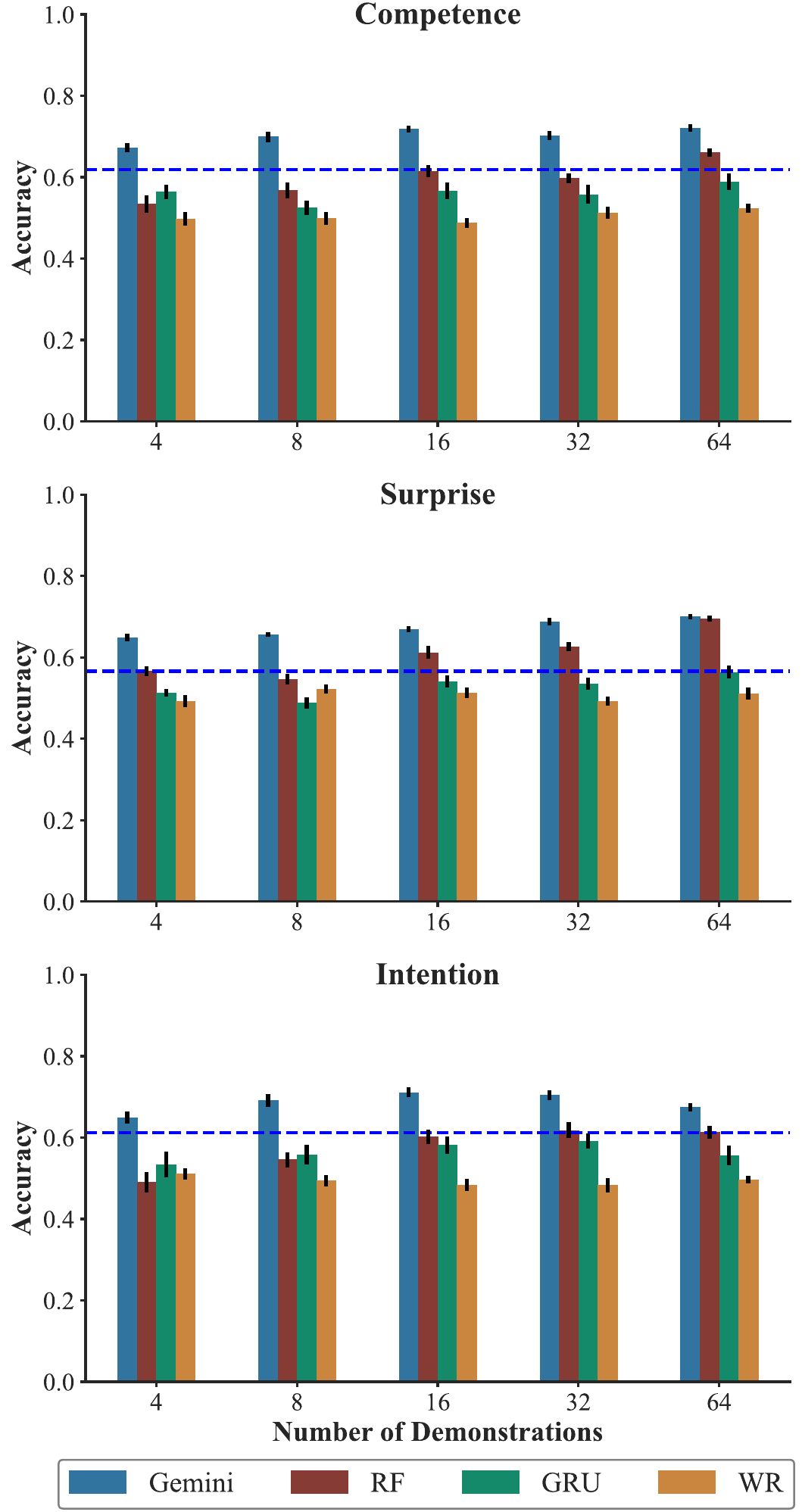}
    \caption{Results for RQ3. Accuracy of Gemini 2.0 Flash No CoT (Gemini), Random Forest (RF), Recurrent Network (GRU), and Weighted Random Sampling (WR) with varying number of demonstrations ($K$). The blue line indicates average accuracy for Gemini 2.0 Flash with a zero-shot prompt.}
    \label{fig:RQ3}
    \vspace{-1em}
\end{figure}

%For RQ2, we investigated whether LLMs can scale up when provided with an increasing number of in-context examples. Based on the RQ1 result, and also considering the scalability and the cost, we performed all the following experiments on Gemini 2.0 Flash with no CoT. And we used two supervised learning baselines, including Random Forest (RF) and a bidirectional Gated Recurrent Unit (GRU) network (GRU) \cite{schuster1997bidirectional}. We also considered a weighted random sampling (WR) baseline from the training set. We considered an increasing number of demonstration examples ranging from 4 to 64, in an order of 2.

\vspace{0.5em}
\noindent
\textbf{Results for Gemini only.} The blue bars and line in Fig.~\ref{fig:RQ3} show average accuracy with an increasing $K$ for Gemini. For each performance dimension, we analyzed accuracy using a linear mixed model with Run ID as a random effect, and Number of Demonstrations ($K \in \{$0, 4, 8, 16, 32, 64$\}$) as a main effects. The analysis showed a significant effect for $K$ on all performance dimensions ($p < .0001$). %Competence ({\color{red}p = XXXX}), Surprise ({\color{red}p = XXXX}), and Intention ({\color{red}p = XXXX}). 
For \textit{Competence}, a Tukey HSD post-hoc test showed that the zero-shot model ($K=0$) led to significantly lower accuracy than all ICL models ($K>0$). Also, ICL with $K \in \{64, 16\}$ had significantly higher accuracy than ICL with $K=4$. For \textit{Surprise}, zero-shot Gemini also led to significantly lower accuracy than ICL. Further, ICL with $K=64$ led to significantly higher accuracy than $K \in \{16, 8, 4\}$. For \textit{Intention}, %the post-hoc test showed that 
ICL with $K \in \{32, 16\}$ led to significantly higher accuracy than $K=4$ and the zero-shot model. Taken together, these results suggest that the LLM benefited from having demonstrations.

\vspace{0.5em}
\noindent
\textbf{Results for Gemini vs Other Models.} 
Fig.~\ref{fig:RQ3} shows results for all the models. %Considering all  performance dimensions, the zero-shot Gemini model was better than random sampling ({\color{red}see the blue line vs. orange bars in Fig.~\ref{fig:RQ3}}). This suggested that the world knowledge encoded by the LLM is beneficial for predicting human perceptions of robots. However, the zero-shot Gemini model underperformed ICL predictions in terms of average performance, suggesting that the demonstrations are useful. Notably, ICL  performance tended to increase with more examples in Fig.~\ref{fig:RQ3}.
We analyzed prediction accuracy on each performance dimension using a linear mixed model, but this time excluded the zero-shot case for which the supervised learning models could not be fit. The linear mixed model considered Run ID as a random effect, Number of Demonstrations ($K \in \{$4, 8, 16, 32, 64$\}$) and Model (\textit{Gemini}, \textit{RF}, \textit{GRU}, \textit{WR}) as a main effects, and the interaction between the Number of Demonstrations and Model. Because of limited space, we focus on discussing the interaction effect, which is the most relevant for RQ3 and was significant for \textit{Competence} (%F(12, 456) = 1.9465$, 
$p = 0.0275$), \textit{Surprise} (%$F(12, 456) = 4.7594$, 
$p < 0.0001$), and \textit{Intention} (%$F(12, 456) = 2.4936$, 
$p = 0.0036$).
%We conducted Tukey HSD post-hoc tests for the effects. %While the average accuracy of Gemini No CoT tended to increase with higher $K$ in Fig. 6, 

The Tukey HSD post-hoc tests for the interaction effect confirmed the superiority of the Gemini model in most cases. For example, for \textit{Competence}, Gemini with $K\in\{64, 32, 16, 8\}$ led to significantly higher accuracy than all other models, except for Gemini with $K=4$ and RF with $K=64$. % which were not significantly different.
In contrast to our prior results for RQ3 considering Gemini only, the post-hoc tests for the interaction effect between the Number of Demonstrations and Model resulted in no significant pairwise differences for Gemini across $K=\{$4, 8, 16, 32, 64$\}$. RF benefited more from an increasing $K$. The $RF$ model with $K=64$ demonstrations led to significantly higher accuracy than $RF$ with $K=\{4,8\}$ across all performance dimensions. %This reinforced our findings  from RQ1 on the limited impact of $K$ on ICL. %, where we considered $K=4$ and $K=64$ only.

\definecolor{mplBlue}{rgb}{0.7176, 0.8431, 0.9490}
\definecolor{mplOrange}{rgb}{0.9843, 0.8902, 0.8353}
\definecolor{mplGreen}{rgb}{0.8000, 0.9250, 0.8000}

\definecolor{mplGreenLight}{rgb}{0.90, 0.97, 0.90}

% Medium green (for 2nd place) – your base pastel green
\definecolor{mplGreenMedium}{rgb}{0.75, 0.90, 0.75}

% Darker green (for 1st place) – clearly deeper but not blocking text
\definecolor{mplGreenDark}{rgb}{0.55, 0.75, 0.55}

\begin{table*}[tb!p]
\fontsize{8.5pt}{10pt}
\selectfont
\centering
\captionof{table}{Results for RQ4. Mean accuracy ($\pm$ std. err.) of Gemini 2.0 Flash with varying numbers of personalized and non-personalized demonstrations, over 25 repetitions. The \colorbox{mplGreenDark}{Best},  \colorbox{mplGreenMedium}{Second}, and \colorbox{mplGreenLight}{Third} average results are highlighted.}
\vspace{-0.1cm}
\begin{tabular}{c|c|c|c|c|c|c}
  \toprule[1.5pt]
  {\color{gray}Row} & 
  \makecell{\textbf{Num. of Non-Personalized} \\ \textbf{Demonstrations ($|D_{\not=}|$)}} & 
  \makecell{\textbf{Num. of Personalized} \\ \textbf{Demonstrations ($|D_{\approx}|$)}} & 
  \makecell{\textbf{Total Number of} \\ \textbf{Demonstrations ($K$)}} & 
  \textbf{Competence} & 
  \textbf{Surprise} & 
  \textbf{Intention} \\
  \midrule
  \addlinespace[0.4em]
  {\color{gray}1} & 0 & 4 & 4 & $0.71 \pm 0.01$ & $0.69 \pm 0.01$ & \cellcolor{mplGreenLight} $0.72 \pm 0.00$ \\
  {\color{gray}2} & 4 & 0 & 4 & $0.67 \pm 0.01$ & $0.65 \pm 0.01$ & $0.65 \pm 0.02$ \\
  {\color{gray}3} & 4 & 4 & 8 & \cellcolor{mplGreenMedium} $0.76 \pm 0.01$ & \cellcolor{mplGreenMedium}  $0.73 \pm 0.01$ & \cellcolor{mplGreenDark} $0.76 \pm 0.01$ \\
  {\color{gray}4} & 8 & 0 & 8 & $0.68 \pm 0.01$ & $0.65 \pm 0.01$ & $0.64 \pm 0.02$ \\
  % 64 & 0 & 64 & $0.72 \pm 0.01$ & \cellcolor{mplGreenLight} $0.70 \pm 0.01$ & $0.67 \pm 0.01$ \\
  {\color{gray}5} & 64 & 4 & 68 & \cellcolor{mplGreenDark} $0.79 \pm 0.01$ & \cellcolor{mplGreenDark} $0.76 \pm 0.00$ & \cellcolor{mplGreenMedium} $0.75 \pm 0.01$ \\
  {\color{gray}6} & 68 & 0 & 68 & \cellcolor{mplGreenLight} $0.72 \pm 0.01$ & \cellcolor{mplGreenLight} $0.70 \pm 0.01$ & $0.69 \pm 0.01$ \\
  \bottomrule[1.25pt]
\end{tabular}
\label{tab:rq3}
\end{table*}

\subsection{ICL with Personalized Demonstrations}

Our last research question was:\\[0.5em]
\noindent
\textit{\textbf{RQ4:} Do personalized examples improve in-context learning?}\\[0.5em]
Because individual factors can influence human perceptions of robots, we examined whether we could improve ICL accuracy with prompts constructed with demonstration examples from the same user who provides the evaluation example (as in Fig.~\ref{fig:prompt_flow}(b)). Following RQ2 and RQ3, we considered only Gemini 2.0 Flash with no CoT for this experiment. %As illustrated on the right side of Fig.~\ref{fig:prompt_flow}, we investigate setups with only personalized examples and setups that blend personalized and general examples, comparing their impact on prediction accuracy.
Also, we considered three values for the total number of ICL demonstrations $K$: 4, 8 and 68 total examples. Specifically, for each $K$ value, we either had:
\begin{itemize}[leftmargin=*,noitemsep,topsep=0pt]
\item[--] $0$ personalized demonstrations (so $|D_{\approx}| = 0$ and $K = |D_{\not=}|$); or 
\item[--] $4$ personalized demonstrations (so $|D_{\approx}| = 4$ and $K = |D_{\not=}| + 4$).
\end{itemize}

\vspace{0.5em}
\noindent
\textbf{Results.} The results are shown in Table~\ref{tab:rq3}. For each performance dimension, we analyzed accuracy using a linear mixed model with Run ID as a random effect, the Number of Personalized Demonstrations ($|D_{\approx}| \in \{0, 4\}$) and the Total Number of Demonstrations ($K \in \{4, 8, 68\}$) as the main effects, and their pairwise interaction. % between $|D_{\approx}|$ and $K$. 

The Number of Personalized Demonstrations ($|D_{\approx}|$) had a significant effect on accuracy on all performance dimensions ($p < 0.0001$). %for Competence (%$F(1, 120) = 72.13$, 
%$p < 0.0001$), Surprise (%$F(1, 120) = 104.58$, 
%$p < 0.0001$), and Intention (%$F(1, 120) = 89.80$, 
%$p < 0.0001$). For all 
The post-hoc test showed that using 4 personalized demonstrations led to significantly higher accuracy than using zero ($|D_{\approx}|=0$). % not using any personalized demonstrations. % competence, Tukey HSD \textit{post hoc} tests showed that using 4 personalized examples ($M = 0.76$, $SE = 0.01$) led to significantly higher accuracy than not using any personalized examples ($M = 0.69$, $SE = 0.01$). For surprise, using 4 personalized examples ($M = 0.73$, $SE = 0.01$) led to significantly higher accuracy than not using any personalized examples ($M = 0.67$, $SE = 0.01$). Also, for predicting intention, ICL using 4 personalized examples ($M = 0.74$, $SE = 0.00$) was significantly more accurate than not using any personalized examples ($M = 0.66$, $SE = 0.01$). 

Additionally, the Total Number of Demonstrations ($K$) had a significant effect on the accuracy. For \textit{Competence} and \textit{Surprise} ($p < 0.0001$), Tukey HSD post-hoc tests indicated that $K=68$ total examples led to significantly higher accuracy than $K=8$, which also led to significantly higher accuracy than $K=4$. For \textit{Intention} ($p = 0.0052$), %the post-hoc tests indicated  
$K=68$  %($M = 0.72$, $SE = 0.01$) 
led to significantly higher accuracy than $K=4$.  % ($M = 0.68$, $SE = 0.01$).
%on the accuracy for the prediction of competence ($F(2, 120) = 23.12$, $p < 0.0001$), surprise ($F(2, 120) = 36.23$, $p < 0.0001$), and intention ($F(2, 120) = 5.50$, $p = 0.0052$). 
%For competence, Tukey HSD \textit{post hoc} tests showed that using 68 total examples ($M = 0.76$, $SE = 0.01$) led to significantly higher accuracy than using 8 total examples ($M = 0.72$, $SE = 0.01$), which is also significantly more accurate than using 4 total examples ($M = 0.69$, $SE = 0.01$). For surprise, using 68 total examples ($M = 0.73$, $SE = 0.01$) led to significantly higher accuracy than using 8 total examples ($M = 0.69$, $SE = 0.01$), which is also significantly more accurate than using 4 total examples ($M = 0.67$, $SE = 0.01$). Also, for predicting intention, ICL using 68 total examples ($M = 0.72$, $SE = 0.01$) led to significantly higher accuracy than using 4 total examples ($M = 0.68$, $SE = 0.01$).

Lastly, we found that the interaction between $|D_{\approx}|$ and $K$ had a significant effect on accuracy for \textit{Surprise} (%$F(2, 120) = 3.77$, 
$p = 0.0258$) and \textit{Intention} (%$F(2, 120) = 3.41$, 
$p = 0.0364$), but not \textit{Competence}. % ($p=0.08$). 
For \textit{Surprise}, using $K=68$ or $K=8$ demonstrations, including $|D_{\approx}| = 4$ examples, led to significantly higher accuracy than the other options in Table~\ref{tab:rq3}. %Further, using only $K = |D_{\approx}| = 4$ personalized demonstrations or $K = 68$ total demonstrations (with no personalized demonstrations) was significantly better than only using $|D_{\not=}| = 8$ or $|D_{\not=}| = 4$ non-personalized demonstrations. 
For \textit{Intention}, using $K=68$ and $|D_{\approx}| = 4$, or using $K=8$ and $|D_{\approx}| = 4$, led to significantly higher accuracy than the other options except for using only $K=|D_{\approx}| = 4$ personalized demonstrations. We conclude that the personalized demonstrations helped ICL performance. % and outperform models that use a higher number of total examples that are not specific to the test user.

\vspace{0.5em}
\noindent
\textbf{Qualitative Analysis.} 
To better understand ICL performance, we manually inspected predictions by Gemini 2.0 Flash No CoT with  $|D_{\not=}|=64$ and $|D_{\approx}|=4$, for a total of 68 demonstrations. As shown in row 5 of Table~\ref{tab:rq3}, this model had highest average accuracy on \textit{Competence} and \textit{Surprise}, and was second best for \textit{Intention}. % by a small margin relative to the top model 

For each perception dimension, we selected three sets of 10 examples from the test set: the 10 examples with the highest prediction accuracy, the 10 with the lowest, and the 10 with accuracy closest to 50\% across the 25 runs of the model. Then, we visualized the navigation episodes and identified recurring patterns that correlated with the model's success, failure, or prediction ambiguity.
%
%We found several patterns. 
Unsurprisingly, the model achieved high accuracy on episodes with consistent robot behaviors, such as steady progress towards the goal or aimless rotation far from it. Low and middle-accuracy predictions mainly stemmed from: 
\begin{description}[wide=0pt, leftmargin=*,font=\normalfont\itshape,itemsep=0.3em]% topsep=0pt,
    \item[1) Semantic ambiguity] of the robot's final state (6/20 examples for \textit{Competence}; 4/20 examples for \textit{Surprise}; and 8/20 examples for \textit{Intention}). We did not provide the LLM a specific threshold for when the robot reached the goal, which made it difficult to gauge how close was close enough to complete the navigation task.
    %    The LLM  lacked the semantic context of ``task completion" (e.g., understanding ``how close is close enough to reaching the goal"), the model could not distinguish an intentional success from an unintentional stall based on the spatial trajectories only.
    \item[2) Transitional ambiguity] from mid-episode behavioral shifts, such as corrective turns or reversals (5/20 examples for \textit{Competence}, and 6/20 examples for \textit{Surprise} and \textit{Intention}). The robot showed both effective and ineffective behavior within an episode. 
    \item[3) Contextual and kinematic ambiguity] in the robot behaviors, like navigating away from the goal or rotating in place (7/20 examples for \textit{Competence}, 6/20 for \textit{Surprise}, and 7/20 for \textit{Intention}). Model uncertainty seemed to stem from subtle variations in the context (e.g., proximity to the goal) or kinematics (e.g., a slow drift vs. a rapid retreat) that were under-sampled in the demonstrations.
\end{description}

\section{Discussion}
\label{sec:limitations}

\noindent
\textbf{Summary of Key Findings.} 
%Our work showed that ICL facilitates inferring human perceptions of a robot with few examples in navigation scenarios, particularly when personalized demonstrations are given to the LLM. Through 
We proposed an In-Context Learning approach for LLMs to infer human perceptions of robot performance. The approach used observations of interactions to predict user evaluations of robot competence, surprisingness, and intent. Our experiments in navigation scenarios showed that our ICL approach not only matches or exceeds the performance of traditional supervised models with a fraction of the data but also outperforms zero-shot LLM predictions. Further, accuracy is enhanced by personalizing in-context examples to the test user.

\vspace{0.3em}
\noindent
\textbf{ICL Limitations.}
Despite its promise, our work also highlighted limitations of ICL. For example, we found mixed results on whether more demonstrations increased prediction accuracy. We suspect the root cause is that with more examples, the context window for the LLM is longer, which can make LLMs struggle~\cite{li2024long}. Also, we investigated choosing demonstrations by random sampling, but perhaps a more thoughtful approach could  help LLMs better leverage more demonstrations~\cite{zhang2022active,liu2022makes}. Importantly, ICL required prompt engineering effort. For example, in early experiments, we found that unintuitive values for the target label could reduce performance, such as using ``surprise = 0'' to indicate surprising behavior. In a qualitative analysis, we also found that semantic ambiguity in what it meant to complete the navigation task in our prompt could lead to erroneous predictions. More systematic experiments are needed to assess the robustness of the ICL approach to prompt variations.

\vspace{0.3em}
\noindent
\textbf{Future Work.}
Limitations of our research also point to future research directions. First, our evaluation focused on a specific robot-following task. More work is needed to validate ICL across more diverse interactions. Second, our observation representation was limited to spatio-temporal features provided as strings to LLMs. Incorporating multimodal cues (e.g., using videos captured from the robot) could increase performance with multi-modal large models. %While we used a consistent prompt structure, future work could also explore the model's robustness to prompt perturbations to ensure reliable performance. 
Ultimately, the most compelling application of this work is robot behavior improvement. A robot could use its predictions of user perceptions as direct feedback to adjust its behavior policy, closing the loop from passive inference to active, socially-aware adaptation.

%\section{Conclusion}

%This paper introduces an in-context learning method that enables Large Language Models to effectively infer human perceptions of robot navigation performance in a highly sample-efficient manner. By converting observed spatio-temporal interactions into structured textual prompts, our method allows LLMs to predict user evaluations of robot competence, surprisingness, and intent. Our experiments show that our in-context learning method not only matches or exceeds the performance of traditional supervised models with a fraction of the data but also that its accuracy is significantly enhanced by personalizing in-context examples to the specific user. As such, our method provides a powerful tool for scalable, user-centered robot evaluation, opening new possibilities for developing robots that can understand and adapt to individual human expectations during real-world interactions.

\section*{Acknowledgements}
% Omitted for blind review.
%To be added upon acceptance.

% Thanks to the National Science Foundation (NSF) under Award No. IIS-2143109
% Thanks to the Google Cloud Platform (GCP) credits program for partially supporting our experiments 

% Thanks to Kenneth.

Thanks to Kenneth Shui for helping with data collection. This work was partially supported by the National Science Foundation (NSF) under Grant No. IIS-2143109, and Google under a Gemini credits grant. Any opinions, findings, and conclusions or recommendations expressed in this paper are those of the author(s) and do not necessarily reflect the views of the NSF or Google.


\bibliographystyle{ACM-Reference-Format}
\balance
\bibliography{references}

\end{document}